\documentclass{article} 

\usepackage{times}
\usepackage{iclr2023_conference}

\usepackage{amsmath,amsfonts,bm}

\def\eqref#1{equation~\ref{#1}}

\def\1{\bm{1}}

\DeclareMathAlphabet{\mathsfit}{\encodingdefault}{\sfdefault}{m}{sl}
\SetMathAlphabet{\mathsfit}{bold}{\encodingdefault}{\sfdefault}{bx}{n}

\usepackage{hyperref}
\usepackage{url}

\usepackage{graphicx}
\usepackage{booktabs}
\usepackage{caption}
\usepackage{floatrow}
\usepackage{paracol}
\usepackage{enumitem}
\usepackage{changepage}
\usepackage{xcolor}
\usepackage{tabularx}
\usepackage{colortbl}

\globalcounter{figure}
\globalcounter{table}

\title{WikiWhy: Answering and Explaining \\ Cause-and-Effect Questions}

\author{Matthew Ho\thanks{Equal contribution} , Aditya Sharma\footnotemark[1] , Justin Chang\footnotemark[1] , \\
\textbf{Michael Saxon, Sharon Levy, Yujie Lu, William Yang Wang}\\
Department of Computer Science, University of California, Santa Barbara, USA\\
\texttt{\{msho}, \texttt{aditya\_sharma}, \texttt{justin\_chang\}@ucsb.edu},\\
\texttt{\{saxon}, \texttt{sharonlevy}, \texttt{yujielu\}@ucsb.edu}, \texttt{william @cs.ucsb.edu}\\
}

\iclrfinalcopy 
\begin{document}

\maketitle


\begin{abstract}

As large language models (LLMs) grow larger and more sophisticated, assessing their ``reasoning'' capabilities in natural language grows more challenging.
Recent question answering (QA) benchmarks that attempt to assess reasoning are often limited by a narrow scope of covered situations and subject matters.
We introduce \textsc{WikiWhy}, a QA dataset built around a novel auxiliary task: explaining why an answer is true in natural language.
\textsc{WikiWhy} contains over 9,000 “why” question-answer-rationale triples, grounded on Wikipedia facts across a diverse set of topics. 
Each rationale is a set of supporting statements connecting the question to the answer.
\textsc{WikiWhy} 
serves as a benchmark for the reasoning capabilities of LLMs because it demands rigorous explicit rationales for each answer to demonstrate the acquisition of implicit commonsense knowledge, which is unlikely to be easily memorized.
GPT-3 baselines achieve only 38.7\% human-evaluated correctness in the end-to-end answer \& explain condition, leaving significant room for future improvements.
\end{abstract}
\section{Introduction}
Error analyses of practical NLP systems in recent history demonstrate that some of the mistakes made by state-of-the-art models would be avoided by basic human intuition \citep{Shuster2022AmIM}, and some of the most challenging tasks for models are the same ones that might be trivial to human children. 
With modern systems' impressive performance on tasks such as grammar correction showing that manipulating language is not the issue, LLMs seem to face a fundamental lack of common sense-- an understanding of everyday phenomena and how they interact with each other and the world at large.
As striking gains in subjective performance on summarization, creative text generation, and apparent language understanding continue to be called into question, the development of strong benchmarks to assess reasoning capabilities for these LLMs grows more important. 

One popular approach to measuring reasoning capability is through performance on question answering (QA) benchmark tasks where direct queries for information act as a straightforward examination of a system's ``understanding.''
Classic QA datasets, however, are primarily concerned with retrieving factoids to answer questions of ``Who'', ``What'', ``When'', and ``Where''.
These questions have been shown to be answerable (with high accuracy) by simple pattern-matching approaches \citep{Wadhwa2018TowardsIR}, thereby limiting their ability to measure the aforementioned reasoning capability.
Looking to maintain the breadth of topics covered while increasing the difficulty of the QA task, researchers introduced multi-hop QA datasets like HotpotQA \citep{yang-etal-2018-hotpotqa}.
While challenging, the task's extra complexity mostly leads to unnatural questions that can be addressed with iterated factoid retrieval and entity resolution, rather than a necessary understanding of how different entities interact. 
Noticeably absent in these prior datasets are ``why'' questions, which prompt for not factoids, but explanations-- reasoning made explicit.

\begin{figure}
    \centering
    \includegraphics[width=1.0  \linewidth]{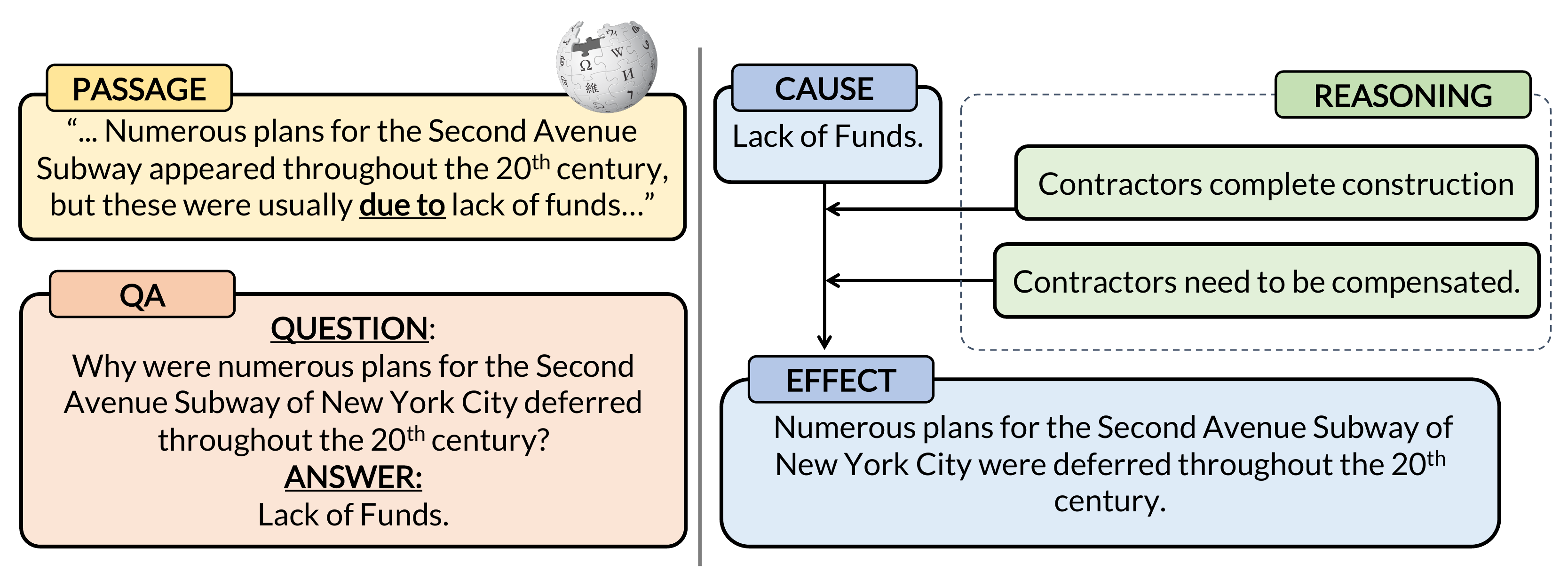}
    \caption{
        A simple example of an entry from \textsc{WikiWhy}; a \textbf{cause} and \textbf{effect} sourced from a Wikipedia \textbf{passage}, a ``why'' \textbf{question} and its \textbf{answer} about this relation, and most importantly \textbf{rationale} that explains why \textbf{cause} leads to \textbf{effect}.
    }
    \label{fig:topexample}
\end{figure}

The task of explanation uses reasoning and produces explicit, interpretable ``thought'' processes. Capitalizing on these properties, this paper introduces \textsc{WikiWhy}, a novel dataset containing ``why'' question-answer pairs. 
Each \textsc{WikiWhy} entry contains a rationale explaining the QA pair's causal relation (\autoref{fig:topexample}), summing to a total of 14,238 explanation elements.
In the context of recent multimodal, self-supervised approaches aiming to capture intuitions unlearnable from text alone \citep{https://doi.org/10.48550/arxiv.2107.10300}, \textsc{WikiWhy} presents an opportunity to investigate a specific kind of information absent in text: implicit commonsense assumptions.
Compared to other QA datasets with rationales, \textsc{WikiWhy} covers a significantly broader range of 11 topics which may prove valuable for developing the skill of applied reasoning on various specific situations. 

Our experiments in explanation generation and human evaluation demonstrate that state-of-the-art generative models struggle with producing satisfying explanations for \textsc{WikiWhy} cause-effect relations.
Our experiments also demonstrate how our proposed task might be used to diagnose a lack of ``understanding'' in certain relations.
Our key contributions are thus:
\begin{itemize}[leftmargin=*]
    \itemsep0em 
    \item We propose explanation \textbf{within} cause-effect relations as a novel problem formulation for exploring LLM reasoning ability.
    \item We create \textsc{WikiWhy}, the first question-answering dataset focusing on reasoning \textbf{within} causal relations, spanning \textbf{11 topics}.
    \item We perform experiments on state-of-the-art, generative models to investigate various settings and establish baseline results with sizable room for improvement.
    \item We introduce idea-level evaluation metrics for free-form text (explanation) generation and a human judgment correlation analysis, demonstrating that (1) reference similarity is strongly correlated with explanation correctness, and (2) the metrics we introduced correlate with this proxy.
\end{itemize}
\section{Related Work}
\textbf{Cause and Effect.  } 
Causality has been a subject of rigorous work in various fields.
In science philosophy, \citet{pearl2009causality} has contributed seminal work relating to causal models, Bayesian networks, and causal strength via interventions and counterfactuals. 
These ideas have even been incorporated into QA tasks through Knowledge Graph approaches, such as filtering spurious latent correlations \citep{cf-kgqa}. 
While our work emphasizes cause-and-effect, we are unconcerned with causal strength as we begin with Wikipedia-grounded relations and are interested in the information encoded into LLMs rather than augmented structures such as knowledge graphs.

\textbf{Multi-hop Question Answering.  }
While datasets such as HotpotQA \citep{yang-etal-2018-hotpotqa} and HybridQA \citep{chen-etal-2020-hybridqa} are instrumental in gauging models' ability to handle multiple sources and modalities, they are focused on iterated factoid retrieval. 
Although chaining multiple facts into a multi-hop answer is useful for products, \textsc{WikiWhy} focuses on \textit{in-filling} rationales to demonstrate reasoning.

\textbf{Visual Question Answering.  }
Vision and language tasks have also intersected with both QA and reasoning.
The Visual Question Answering (VQA) dataset \citep{agrawal-vqa} prompts textual answers to questions about images. However, the caption-based generation leads to surface-level questions that require little reasoning ability, and the multiple-choice output format precludes explicit reasoning. 
The vision-based Sherlock dataset \citep{hessel-sherlock} is much closer to our work, focusing on abductive reasoning (working backward from a consequence).
Setting aside modality differences, \textsc{WikiWhy} requires deeper reasoning with its multi-hop explanations.

\begin{table}[t!]
    \centering
    \caption{A comparison of \textsc{WikiWhy} with previous QA datasets relating to explanation}
    \begin{tabular}{|l|r|c|c|r|c|}
        \specialrule{.15em}{.05em}{.05em}
        \textbf{Dataset} & Size & Answer Type & Explanation Type & Topics & Source\\
        \specialrule{.075em}{.05em}{.05em} 
        CoS-E$^1$ & 9,500 & MCQ & 1-step & 1 & ConceptNet\\
        eQASC$^2$ & 9,980 & MCQ & 2-step & 1 & WorldTree\\
        CausalQA$^3$ & 24,000 & Short & None & 1 & Yahoo Finance \\
        EntailmentBank$^4$  & 1,840 & Short & Tree & 1 & WorldTree \\
        \specialrule{.05em}{.05em}{.05em} 
        \textsc{WikiWhy} & \textbf{9,406} & \textbf{Short} & \textbf{Set/Chain} & \textbf{11} & \textbf{Wikipedia}\\
        \specialrule{.15em}{.05em}{.05em}
    \end{tabular}
    $^1$\citep{rajani-etal-2019-explain}, $^2$\citep{jhamtani-clark-2020-learning}, $^3$\citep{yang2022towards}, $^4$\citep{dalvi-etal-2021-explaining}
    \label{tab:compare}
\end{table}

\textbf{Explainable QA.  }
One previous approach to building explanation resources collects direct answers to ``why'' questions.
TellMeWhy \citep{Lal_2021} features question-answer pairs tied to short story narrative contexts. 
The dataset skips step-wise explanations, prioritizing reading comprehension instead.
On the other hand, ELI5 \citep{eli5LQA} dives deep into reasoning with long-form, detailed explanations.
However, the open-endedness (compared to explaining a specific cause-effect relation) complicates evaluating candidate responses.

Another line of QA work emphasizes a rationale component as support for answer predictions.
Datasets like CoS-E \citep{rajani-etal-2019-explain}, eQASC\citep{jhamtani-clark-2020-learning}, and EntailmentBank \citep{dalvi-etal-2021-explaining} focus on explanation and reasoning much like \textsc{WikiWhy}, albeit with significant differences (\autoref{tab:compare}).
CoS-E's explanations for CommonsenseQA \citep{talmor-etal-2019-commonsenseqa} mark an important first step, but the commonsense explanations have limited depth, often requiring a single hop of reasoning.
eQASC and EntailmentBank feature richer explanations with more complex structure, tightly focusing on grade school level science facts.
Regarding structure, fixed-length rationale in CoS-E, eQASC, FEVER \citep{thorne-etal-2018-fever}, and e-SNLI \citep{eSNLI} capture less granularity, while entailment trees accept limitations in scale and naturalness in exchange for complete ordering information.
Previous datasets tend towards retrieval tasks with eQASC's corpus of all rationale sentences and EntailmentBank's collection of root causes.
Retrieval enables simple evaluation, at the cost of decreased difficulty, the possibility for exploiting spurious artifacts, and reduced debugging opportunity.
\section{Background}
\subsection{Why focus on ``Why'' Questions?}
``Why'' questions are underrepresented in other QA datasets. Users tend to ask straightforward questions that use words like ``who'', ``what'', ``when'' or ``where.'' Questions of this more common form have simple answers that state standalone facts which may be elaborated but do not require explanation. Consider the pair, ``Q: Where do the Tigris and Euphrates rivers meet? A: The Persian Gulf.'' The answer is straightforward.

\begin{figure}
    \centering
    \includegraphics[width=1.0\linewidth]{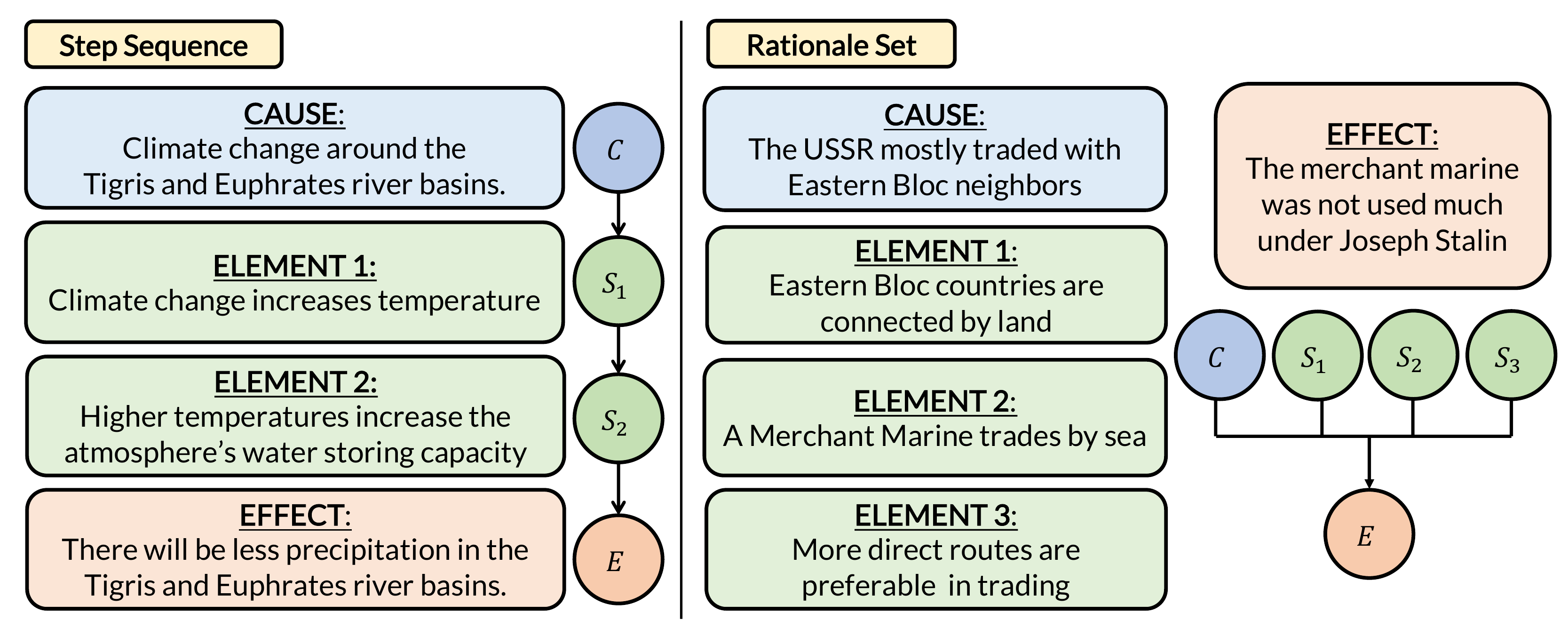}
    \caption{Explanation topologies in \textsc{WikiWhy} mainly vary between a sequence of intermediate conclusions (chain-like) and a set of rationale that combine with the original cause to entail the final effect.}
    \label{figure:chaintree}
\end{figure}

In contrast, a ``why'' QA-pair encodes a cause-effect relation. 
Take, for example, ``Q: Why are precipitation levels falling in the Tigris and Euphrates river basin? A: Climate Change.'' This pair encodes the causal relation ``Climate change is reducing the amount of precipitation in the Tigris and Euphrates river basin'' (\autoref{figure:chaintree}).
The answer to a ``why''-question is an explanation itself (climate change explains reduced precipitation), but we can take it a step further and ask ``why'' \emph{again} to request the understanding or intuition of this process.
While there are some processes at the edge of human understanding or taken as axioms, we assert that there are valid explanations for most processes due to the layered nature of human understanding.
This extra step is especially worth taking since it allows \textsc{WikiWhy} to not only test if a model ``knows'' that ``climate change causes reduced precipitation`` but also if it ``understands'' the underlying mechanics of why that is the case.

\subsection{Task Formulation}
Formally defined in \S\ref{sec:exp}, we propose a \textbf{generative} explanation task. 
Previous works have made strides in assessing reasoning through multiple choice \citep{lu-learntoexplain}, retrieval \citep{asai-wikipediaretrieval}, and partial generation \citep{dalvi-etal-2021-explaining}. 
While these works are undoubtedly crucial towards the end goal of understanding and reasoning, their task formulations have some drawbacks. 
Referring back to education, studies on human students have shown that multiple choice questions ``obscure nuance in student thinking'' \citep{Hubbard2017HowQT}. 
Likewise, a selection decision can be correct for retriever systems but for the wrong reasons.
Augmenting multi-hop factoid questions with an additional task of selecting the relevant supporting facts from the context passage, \citet{Inoue2020R4CAB} emphasizes that interpretability is lost in the absence of explanation.
Furthermore, text generation to combine existing ideas is arguably a different task than generating from scratch. 
The field of psychology defines recall (mental retrieval of information) as a distinct process from recognition (mental familiarity with the cue) \citep{Mohr1989RecallAR}. 
Neural nets' biological inspiration suggests that there might be a similar difference between cue-aided retrieval and freeform generation. 
In the context of NLP, we are interested in the implicit understandings and assumptions embedded in LLMs and hypothesize that an entirely generative approach is most conducive to this study.


\subsection{Explanation Structure}
\label{sec:topologies}
Explanations come in various structures, as seen in the typology defined by \citet{Ribeiro2022EntailmentTE}. 
Shown in \autoref{figure:chaintree}, our work focuses on a subset of said typology. 
\textsc{WikiWhy} includes two structures that explain cause-and-effect relations: (1) multi-hop step sequences and (2) rationale sets. While the chain structure adds intermediate conclusions between cause and effect, rationale sets contain elements that support the relation from without.
The rationale set topology acts as our general, catch-all case that other structures can be condensed to.
Since our data collection procedure promotes a stepwise, ordered approach, we also consider the sequential topology to respect the structure exhibited in applicable explanations.
We forego the unstructured approach as even limited structure helps bring freeform generated text evaluation within reach. 
Finally, we opt against pursuing the most complex entailment tree organization to maintain naturalness and facilitate crowdsourcing scalability.
\section{Dataset}
\subsection{Data Collection}
The objective of \textsc{WikiWhy} is to present a high-quality, challenging dataset of QA pairs with corresponding causes, effects, and explanations. We developed an extensive data collection and validation pipeline around Amazon Mechanical Turk, depicted in \autoref{figure:pipeline} (appendix). For each stage involving crowdsourced annotations, we perform rigorous worker-level quality control to ensure the dataset's quality. The exact procedures are detailed in \autoref{appendix:validation} in the appendix.

\paragraph{Preprocessing.}
We begin with English Wikipedia's corpus of ``Good Articles,"\footnote{\href{https://en.wikipedia.org/wiki/Wikipedia:Good_articles/all}{https://en.wikipedia.org/wiki/Wikipedia:Good\_articles/all}}, whose strict criteria of verifiability and neutrality (among others) ensure that \textsc{WikiWhy} does not evaluate models on misinformation or opinionated views.
From these articles, we extract passages containing causal relations using causal connectives.
We selected a list of causal keywords (Appendix, \S \autoref{appendix:collection}) from a more extensive set of causal connectives as their presence in a passage guarantees the existence of a cause and effect relation---some excluded connectives such as ``since'' or ``as'' are highly prevalent but are not necessarily causal. 
The presence of a causal word pattern on its own is a very simple heuristic---in the subsequent collection steps, we hired crowdworkers to ensure the quality of each sample.

\paragraph{QA Synthesis (Stage 1).}
Randomly sampled preprocessed Wikipedia passages containing potential causal statements were shown to qualified Amazon Mechanical Turk (MTurk) workers (see ethics statement for details), who were tasked with extracting the highlighted causal relation from the passage and re-framing it as a ``why'' question when possible. 
While automatic cause-effect relation extraction has seen recent progress \citep{yang2022towards}, this human intelligence task (HIT) remains vital for two reasons. 
First, we find that quality in cause-effect is crucial for meaningful and valid explanations in the following stage. 
More importantly, we depend on human annotators to add sufficient context to the text of the cause, effect, and question to disambiguate them. 
This enables the question and cause-effect relation to be presented to models without the context we prepared (e.g., ``Why was the river diverted?'' is unanswerable without additional context). This feature is key to enabling \textsc{WikiWhy} to assess the information and ideas within LLMs as opposed to whatever may be present in the context.

\paragraph{Explanation Synthesis (Stage 2).}
After verifying the quality of the examples, we prompt crowd workers to explain cause-effect pairs from stage 1. 
To encourage structured explanation, we supply an interface that allows sentences or ideas to be entered one at a time in separate fields.
Though the input pairs should be context-independent, we provide the original passage as an aid for understanding the topic.
Furthermore, we provide the link to the source article to encourage explanations leveraging topic-specific information in addition to commonsense knowledge.

\subsection{Dataset Description}
\textbf{Entry Contents. }
In addition to the main fields of the question, answer, and explanation, each dataset entry contains the underlying relation's cause and effect, the passage the question was extracted from, the article the passage is from, and Wikipedia's topic categorization for that article.

\textbf{Topic Diversity. }
\textsc{WikiWhy} improves upon other datasets due to its ability to examine reasoning proficiency across a broader range of concepts (\autoref{tab:genre_diversity} in Appendix contains examples from the six most frequent topics).

\textbf{Rationale. }
The statistics for the reasoning component are shown in Appendix \autoref{tab:summary_stats}. 
On average, each rationale contains \textbf{1.5137} elements. 
\autoref{fig:rationale_length} (Appendix) shows a histogram of rationale length by sentence count. 
\textsc{WikiWhy} includes a range of rationale lengths, with more than one-third of examples (36\%) containing two or more reasoning steps.
\section{Experiments}
\label{sec:exp}
\subsection{Experimental Settings and Models}
\paragraph{Task Notation}
Let $C$ be a cause clause; $E$ be an effect clause corresponding to $C$; $Q$ be a question equivalent to ``Why is it the case that $E$?''; $A$ be the answer to $Q$ \footnote{Note that $Q$ is a query that provides $E$ and is correctly answered by $C$, $C = A$.}; $X$ be the explanation = $(S_1, S_2,\ldots, S_k)$ where $S_i$ is a sentence such that:
\[C \land S_1 \land S_2 \land \ldots \land S_k \vdash E\]
\paragraph{Task 1: Question Answering (QA). Input = $Q$, Output = $A$.}
For thoroughness, we confirm high performance on Task 1 (Standard QA) in the open-book setting.
For this set of experiments, we use the classic approach of breaking the task into separate retrieval and reading comprehension phases.
We experiment with BM25 \citep{robertson2009probabilistic} and Dense Passage Retriever (DPR) \citep{dpr} as our document retriever, using their Pyserini implementations \citep{pyserini}. 
Using the Natural Questions \citep{kwiatkowski-etal-2019-natural} encoder, as in the original DPR paper, we build custom indices around segments from the subset of Wikipedia Articles shown to workers at collection time.
For reading comprehension, we experimented with RoBERTa \citep{roberta} and Big Bird \citep{Zaheer2020BigBT} QA models. We also fine-tune a Fusion-in-Decoder (FiD) \citep{fid} model (80-10-10 split; default configurations), hypothesizing the decode-time combination of ideas could better model cause-effect relations. 

The performance was unsurprisingly high, with BM25 achieving a high Top-1 Accuracy score of 0.810 in retrieval and FiD reaching a mean BERT-f1 of 0.78 (\autoref{tab:retrieval} in Appendix).
While retrieving the appropriate Wikipedia passage relating to some topic is straightforward, we found that producing an explanation of comparable quality to our gold rationales was difficult for the models we tested.

\paragraph{Task 2: Explanation Only (EO). Input = $(C, E)$, Output = $X$.}
First, we examine task 2: generating an explanation given an initial cause-effect pair. 
Given their stronger zero-shot generalization \citep{Wang2022WhatLM}, we choose decoder-only models for our baselines. 
In this vein, we investigate the few-shot abilities of GPT-3 \citep{brown2020language} with OpenAI's most capable model, DaVinci-002, at otherwise default settings. 
To better coax out the intermediates between cause and effect, we conduct prompt engineering over \citet{Wei2022ChainOT}'s Chain of Thought method. 
Our exemplars are shown in \autoref{fig:exemplars}.

We also make use of \textsc{WikiWhy}'s scale for fine-tuning GPT-2 \citep{Radford2019LanguageMA}. 
In this set of experiments, we attempt to balance improving GPT-2's understanding of the task's structure while preserving the model's ``intuitions'' for examination. 
We train GPT-2 for ten epochs using the training split ($\approx$80\% of the data) and Adam \citep{kingma2014adam} optimizer with standard hyperparameters ($\text{learning rate} = .001, \beta_1 = .9, \beta_2 = .999, \epsilon = 1\text{e-}8, \text{decay}=0$). 
For this tuned model we introduce special delimiter tokens \texttt{<cause>}, \texttt{<effect>}, and \texttt{<explanation>} in addition to the beginning and end tokens \texttt{<bos>} and \texttt{<eos>}.
To support the delimiters and help the model distinguish the segments, we add token type embeddings (marking cause, effect, and explanation) as part of the preprocessing phase.
At decoding time, we experiment with multiple temperatures.

\paragraph{Task 3: Answer and Explanation (A\&E). Input = $Q$, Output = $(A, X)$.}
To investigate the performance of jointly predicting an answer and explanation given only a ``why'' question, we carry forward with our best performing baseline from the \textbf{EO} task, chain-of-thought prompted GPT-3. 
The first setting in this experiment set tasks a single model with the full end-to-end procedure. 
Once again, we utilize Chain-of-Thought prompting, albeit with a modified prompt that also requests an answer to handle the different input format.
Considering the impressive performance of existing IR techniques on the \textbf{QA} task described above, we also study an additional setting incorporating the \textbf{QA} task.
In the ``pipeline'' setting, the explainer model still lacks access to the ideal answer (the explanation's starting point) but benefits from a reader model's access to the original context.
Here we combine our strongest performing approaches to the \textbf{QA} and \textbf{EO} tasks to make a 3-step pipeline of retrieval (BM25), reading (FiD), and explanation (GPT-3).
\subsection{Automatic Evaluation Metrics}
While the still developing area of text generation has measures and proxies for similarity that help with simple sequences, comparing reasoning sequences or rationale sets requires more involved measures. 
With the two topologies introduced in \S \ref{sec:topologies} in mind, we propose two related metrics, unordered and ordered, to handle sets and sequences, respectively.

\paragraph{Unordered Evaluation.}
This first approach compares the ideas contained in the predictions and references. First, we split predicted and reference explanations into ``ideas'' or ``steps'' by sentence. 
We then compute a matrix of pairwise similarity scores before using a threshold to classify ``matches''. 
Since a single prediction sentence may contain multiple reference ideas, we keep separate counts of precise prediction steps and covered reference steps. 
These counts are then micro-averaged for the test set's overall precision, recall, and F1 scores. 

\begin{figure}[tp]
    \centering
    \includegraphics[width=1\linewidth]{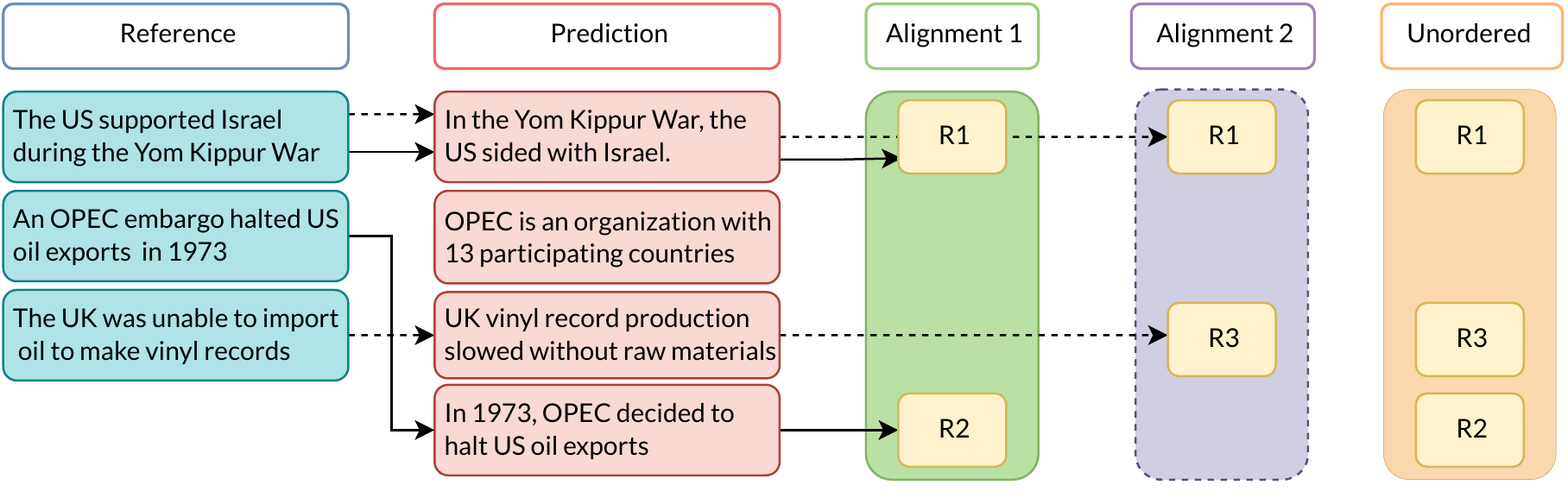}
    \caption{
        Alignment example for sentence-level metrics. Ordered evaluation uses the longest common subsequence as shown by alignment 1 and 2. The final alignment's length is used to compute F-score metrics.
    }
    \label{fig:ordered_alignment}
\end{figure}

\paragraph{Ordered Evaluation.}
To respect the structure of multi-hop explanations, we penalize incorrectly ordered explanations. 
Here, we use the previously generated pairwise score matrix and its alignments to generate all possible assignments of prediction sequence elements to reference elements. 
As demonstrated in \autoref{fig:ordered_alignment}, we compute the length of the longest common subsequence (LCS) between a prediction alignment against the reference labels for each candidate assignment.
This length becomes the count of correctly incorporated structural elements-- true positives.
Note that under this scheme, the repeated ideas in the prediction are discounted by the LCS-style alignment process.

\paragraph{Metric Validity.}
To understand the usefulness of our constructed metrics, we compare them against human judgements. 
A panel of 3 undergraduate students compared pairs of predictions and references on two binary scales: (\textbf{1. Similarity}) ``Is the prediction similar to the reference?'' and (\textbf{2. Correctness}) ``Is the prediction a valid or correct explanation of the cause-effect pair?'' 
Summing the panelist scores for each pair, we found a strong correlation ($r = 0.82$) between the similarity and correctness judgement.
This validates comparison with \textsc{WikiWhy} gold explanations as a useful proxy for explanation quality. Our proposed sentence-level processing incorporates the intuitions of checking for completeness with recall and penalizing over-explanation with precision.

Further, we use a single-explanation version of F-score to compare this proposed automatic metric with human judgement (the proposed F-score measures aggregate through the whole dataset). With this variation, we find a modest correlation ($r = 0.35$) between ordered F1 and similarity, among other weaker correlations.

Besides supporting our proposed methods, this correlation analysis also enabled a data-driven approach to calibrating our similarity metric and match criteria.
For each similarity metric, we selected a starting point through manual inspection of prediction-reference-similarity triples (which threshold value divides ``genuine'' from mistaken similarity) and used correlation for refinement.  
After trials with BLEURT \citep{Sellam2020BLEURTLR} and BERTScore \citep{Zhang2020BERTScoreET}, different underlying models and different match thresholds, we selected BERTScore using a large DeBERTa \citep{He2021DeBERTaDB} model (\texttt{microsoft/deberta-xlarge-mnli}) at a threshold of 0.64.

\subsection{Human Evaluation}
Recent studies by \citet{goyal-gpt-eval} show that automatic metrics may not reliably evaluate results produced by models with few-shot capabilities like GPT-3. 
In light of this, we supplement our automatic evaluation with an additional human evaluation. We first evaluate each setting in each experiment using the binary correctness scale (see criteria definition below).
Following this evaluation, we select the highest scoring explanations for each set of experiments for additional fine-grained evaluation. 
For each human evaluation task, we present a panel of three undergraduate students a random sample of 50 entries from each setting and the following binary True/False criteria guidelines:
\begin{itemize}[leftmargin=*]
    \item \textbf{Correctness}: Mark true if and only if the explanation is both complete and satisfying.
    \item \textbf{Concision}: Mark true if the explanation says everything it needs to say and nothing more. Mark false if extra information is included. 
    \item \textbf{Fluency}: Is the explanation writing fluent? Mark false if there are any mechanical mistakes.
    \item \textbf{Validity}: Does the explanation make logical sense? Ignore whether or not the explanation successfully explains the cause/effect relation. Mark false if the explanation contains any illogical or untrue conclusions.
    \item \textbf{Win/Tie/Lose}: Compare the generated explanation against the provided reference (\textsc{WikiWhy} gold explanation). Mark Win if you prefer the generated explanation, Tie if you have no preference, and Lose if you prefer the reference explanation.
\end{itemize}

\subsection{Results}
\paragraph{Fine-Grained Human Evaluation.}
With our human evaluation experiments, we find significant room for improvement across the board. 
First and foremost, our strongest baseline, GPT-3 with greedy decoding, produced explanations judged to be satisfactory only 66\% of the time (\autoref{tab:human_eval}).
Moreover, these explanations were judged to be worse than the gold reference 58\% of the time.
These results from our strongest baseline leave plenty of room to improve upon and motivate future work on this reasoning task.

\paragraph{Decoding.}
Our experiments show increased performance with lower temperature sampling and best results with greedy decoding  (\autoref{tab:gpt2}).
This aligns with existing notions of higher temperatures better suiting ``creative,'' open-ended tasks as opposed to more grounded ones.
Explaining, as we hypothesize, relies more on the embedded assumptions in a generative model rather than the unlikely associations made more likely at higher temperatures.

\paragraph{Model Differences.}
We find that GPT-3 significantly outperforms GPT-2. 
Comparing GPT-3's output against its predecessor's strongest setting shows increases in both ordered and unordered F1 scores by over 50\%. 
Despite benefiting from fine-tuning and additional structural support from token type embeddings, GPT-2's explanations are lacking compared to GPT-3's few-shot explanations using only 4 exemplars. 
We find that GPT-2's statements are often not only incomplete/unsatisfying for explaining the cause-effect relation at hand but also simply invalid. 
94\% of GPT-2's statements were deemed worse than \textsc{WikiWhy}'s gold references. 
The only area GPT-2 outperformed GPT-3 was in concision, however this is more a demerit of GPT-3 rather than a merit of GPT-2. 
We found that GPT-3 tended to occasionally add unnecessary detail to its explanations, often defining one of the entities in the prompt.

\begin{table}[t]
  \centering
    \caption{Baseline Performance on Explanation Tasks (EO = Explanation-Only, A\&E: Answer and Explanation). For Task 3, the Single Model setting has the generative model complete the end-to-end task in a single pass. The Pipeline setting allows each stage to be handled separately (QA is handled by BM25+FiD and explanation is done by GPT-3). Human evaluation was done with on a binary scale (correct/incorrect) and we report the proportion of correct evaluations.} 
    \begin{tabular}{lcccccccc}
     \specialrule{.15em}{.05em}{.05em}
      \textbf{Experiments} & \multicolumn{3}{c}{\textbf{Unordered}} & & \multicolumn{3}{c}{\textbf{Ordered}} & \textbf{Human}\\
    \cmidrule{2-4} \cmidrule{6-8}
     & Precision & Recall & BERT-f1 & & Precision & Recall & BERT-f1 & Correct \\
    \specialrule{.075em}{.05em}{.05em}
    \textbf{Task 2: EO} \\
    ~~GPT-2 \\
    ~~~~~\emph{Greedy} & 0.249 & 0.196 & 0.220 & & 0.239 & 0.179 & 0.204 & \textbf{0.100} \\
    ~~~~~\emph{$T=0.50$} & 0.218 & 0.164 & 0.188 & & 0.194 & 0.146 & 0.166 & 0.065\\
    ~~~~~\emph{$T=1.00$} & 0.072 & 0.056 & 0.063 & & 0.071 & 0.054 & 0.062 & 0.064\\
    ~~GPT-3 \\
    ~~~~~\emph{Greedy} & 0.347 & 0.388 & \textbf{0.366} & & 0.307 & 0.355 & \textbf{0.329} & \textbf{0.660} \\
    ~~~~~\emph{$T=1.00$} & 0.326 & 0.356 & 0.340 & & 0.291 & 0.328 & 0.308 & 0.481 \\
    \specialrule{.075em}{.05em}{.05em}
    \textbf{Task 3: A\&E} \\
    ~~GPT-3 \\
    ~~~~~\emph{Single-Model} & 0.092 & 0.095 & 0.094 & & 0.082 & 0.092 & 0.087 & 0.140 \\
    ~~~~~\emph{Pipeline} & 0.229 & 0.233 & \textbf{0.231} & & 0.211 & 0.220 & \textbf{0.215} & \textbf{0.387} \\
     \specialrule{.15em}{.05em}{.05em}
    \end{tabular}
    \label{tab:gpt2}
\end{table}

\begin{table}[t]
    \caption{Human evaluation. Overall correctness is marked on a binary scale-- an explanation is complete and satisfying or not. Concision penalizes for repeated or unnecessary information, fluency evaluates grammar, and validity measures if the generated sequence makes logical sense regardless if it correctly explains the relation. For Win/Lose/Tie, annotators compared the generations against \textsc{WikiWhy}'s gold references.}
    \centering
    \begin{tabular}{lc|ccc|ccc}
    \specialrule{.15em}{.05em}{.05em}
    \textbf{Setting} & 
    \multicolumn{7}{c} {\textbf{Fine Grained Human Evaluation}} \\
    \cmidrule{2-8}
    & Correctness & Concision & Fluency & Validity & Win ($\uparrow$) & Tie & Lose ($\downarrow$)\\
    \specialrule{.075em}{.05em}{.05em}
    \textbf{GPT-2: EO} & 0.100 & 0.880 &  0.860 & 0.520 & 0.040 & 0.040 & 0.920\\
    \textbf{GPT-3: EO} & \textbf{0.660} & 0.680 & 1.00 & 0.960 & 0.080 & 0.360 & 0.580 \\
    \textbf{GPT-3: A\&E} &  0.140 & 0.680 & 0.900 & 0.720 & 0.080 & 0.100 & 0.820 \\
    \specialrule{.15em}{.05em}{.05em}
    \end{tabular}
    \label{tab:human_eval}
\end{table}

\paragraph{Answer \& Explanation.}
On the A\&E task, we find results that align cleanly with preconceived intuitions.
Our baseline model is able to better handle explanations from points A to B when A is fixed and provided.
Requiring the same procedures to generate more output creates more variance as incorrect or alternative starting points mislead the remaining generation.
The ``pipeline'' setting strengthens this trend, as the better-informed answer generation allows for a higher quality explanation.
This setting, simulating a three-step process with different models handling each step, demonstrates an intermediate performance between having the oracle-provided answer and requiring the explainer to manage the entire process.
Under these settings, where the model's input excludes the correct answer (the cause), the ``validity'' criteria of our human evaluation is especially interesting.
While the majority of the end-to-end setting's explanations were marked incorrect or unsatisfying, a notable proportion was still marked as having a valid chain of reasoning. 
This suggests that a significant portion of this setting's difficulty lies in the generation of the initial correct answer. 

\paragraph{Explanation Failure.}
A typical error pattern observed in GPT-3's predictions is repeating the cause-effect relation. 
To explain why \texttt{[A]} leads to \texttt{[B]}, GPT-3 might only write ``\texttt{[B]} because \texttt{[A]}'' or another semantically equivalent formulation.
This pattern may be explainable with a fine-tuned baseline where annotation errors of the same kind might have slipped into the training set, but GPT-3 was prompted with hand-picked exemplars with no such mistakes.
Furthermore, we observe successful explanations on some inputs we expect to be more difficult alongside errors on relatively less challenging inputs. 
These observations, together with the consistently high fluency scores showing syntactic competence, seem to indicate a reasoning failure as opposed to a systematic ``misunderstanding'' of the task at hand.
Per the original goal of better understanding what and how LLMs ``understand'' the world, this might indicate a gap in commonsense: that GPT simply memorized the fact that \texttt{[A]} leads to \texttt{[B]}.
\section{Conclusion}
With this paper, we release \textsc{WikiWhy}, a Question-Answering dataset enabling the analysis and improvement of LLMs' reasoning capability.
We propose explanation \textbf{between} grounded cause-effect pairs to distinguish memorization of the relation from a genuine understanding of the underlying mechanics.
Compared to related works on explainable QA, our explanation format finds a natural middle ground that balances \textbf{complexity} and \textbf{depth}, allowing our crowdsourcing methods to produce thought-provoking examples while being highly scalable. 
We exploit this scalability to cover topics previously overlooked by other explanation datasets and demonstrate our proposed task to be difficult with strong baselines (our experiments feature models failing to produce satisfying explanations even under ideal conditions). 
Finally, we motivate the development of new automatic metrics that are better able to handle the complexities of generated reasoning.

\section*{Ethics Statement}
\label{sec:ethics}
For data collection, our listing required workers to have a high HIT approval rating ($\geq96\%$) and be located in English speaking regions (Australia, Canada, New Zealand, the United Kingdom, and the United States). The average hourly pay is 12.00 dollars, which exceeds the income requirements proposed in the human subjects research protocols. The project is classified as exempt status for IRB. Our interfaces include notices that we are collecting information for dataset creation, consent forms, and a link for inquiries and concerns. Our MTurk interfaces are displayed in the \S \ref{sec:appendix}. Due to the experimental nature, limited production applicability, and relatively small dataset scale, we believe the potential for misuse or harm is negligible. 

\section*{Reproducibility Statement}
We release our codebase containing the model tuning procedures, settings, few-shot prompts, and evaluation script under supplementary materials.

\section*{Acknowledgements}
This work is supported by the Amazon AWS AI/ML Research Award and AWS Cloud Credit for Research.

\bibliography{iclr2023/iclr2023_conference}

\begin{thebibliography}{44}
\providecommand{\natexlab}[1]{#1}
\providecommand{\url}[1]{\texttt{#1}}
\expandafter\ifx\csname urlstyle\endcsname\relax
  \providecommand{\doi}[1]{doi: #1}\else
  \providecommand{\doi}{doi: \begingroup \urlstyle{rm}\Url}\fi

\bibitem[Agrawal et~al.(2015)Agrawal, Lu, Antol, Mitchell, Zitnick, Batra, and
  Parikh]{agrawal-vqa}
Aishwarya Agrawal, Jiasen Lu, Stanislaw Antol, Margaret Mitchell, C.~Lawrence
  Zitnick, Dhruv Batra, and Devi Parikh.
\newblock Vqa: Visual question answering, 2015.
\newblock URL \url{https://arxiv.org/abs/1505.00468}.

\bibitem[Asai et~al.(2019)Asai, Hashimoto, Hajishirzi, Socher, and
  Xiong]{asai-wikipediaretrieval}
Akari Asai, Kazuma Hashimoto, Hannaneh Hajishirzi, Richard Socher, and Caiming
  Xiong.
\newblock Learning to retrieve reasoning paths over wikipedia graph for
  question answering, 2019.
\newblock URL \url{https://arxiv.org/abs/1911.10470}.

\bibitem[Brown et~al.(2020)Brown, Mann, Ryder, Subbiah, Kaplan, Dhariwal,
  Neelakantan, Shyam, Sastry, Askell, et~al.]{brown2020language}
Tom Brown, Benjamin Mann, Nick Ryder, Melanie Subbiah, Jared~D Kaplan, Prafulla
  Dhariwal, Arvind Neelakantan, Pranav Shyam, Girish Sastry, Amanda Askell,
  et~al.
\newblock Language models are few-shot learners.
\newblock \emph{Advances in neural information processing systems},
  33:\penalty0 1877--1901, 2020.

\bibitem[Camburu et~al.(2018)Camburu, Rocktäschel, Lukasiewicz, and
  Blunsom]{eSNLI}
Oana-Maria Camburu, Tim Rocktäschel, Thomas Lukasiewicz, and Phil Blunsom.
\newblock e-snli: Natural language inference with natural language
  explanations, 2018.
\newblock URL \url{https://arxiv.org/abs/1812.01193}.

\bibitem[Chadha \& Jain(2021)Chadha and
  Jain]{https://doi.org/10.48550/arxiv.2107.10300}
Aman Chadha and Vinija Jain.
\newblock ireason: Multimodal commonsense reasoning using videos and natural
  language with interpretability, 2021.
\newblock URL \url{https://arxiv.org/abs/2107.10300}.

\bibitem[Chen et~al.(2020)Chen, Zha, Chen, Xiong, Wang, and
  Wang]{chen-etal-2020-hybridqa}
Wenhu Chen, Hanwen Zha, Zhiyu Chen, Wenhan Xiong, Hong Wang, and William~Yang
  Wang.
\newblock {H}ybrid{QA}: A dataset of multi-hop question answering over tabular
  and textual data.
\newblock In \emph{Findings of the Association for Computational Linguistics:
  EMNLP 2020}, pp.\  1026--1036, Online, November 2020. Association for
  Computational Linguistics.
\newblock \doi{10.18653/v1/2020.findings-emnlp.91}.
\newblock URL \url{https://aclanthology.org/2020.findings-emnlp.91}.

\bibitem[Clark et~al.(2019)Clark, Celikyilmaz, and
  Smith]{clark-etal-2019-sentence}
Elizabeth Clark, Asli Celikyilmaz, and Noah~A. Smith.
\newblock Sentence mover{'}s similarity: Automatic evaluation for
  multi-sentence texts.
\newblock In \emph{Proceedings of the 57th Annual Meeting of the Association
  for Computational Linguistics}, pp.\  2748--2760, Florence, Italy, July 2019.
  Association for Computational Linguistics.
\newblock \doi{10.18653/v1/P19-1264}.
\newblock URL \url{https://aclanthology.org/P19-1264}.

\bibitem[Dalvi et~al.(2021)Dalvi, Jansen, Tafjord, Xie, Smith, Pipatanangkura,
  and Clark]{dalvi-etal-2021-explaining}
Bhavana Dalvi, Peter Jansen, Oyvind Tafjord, Zhengnan Xie, Hannah Smith,
  Leighanna Pipatanangkura, and Peter Clark.
\newblock Explaining answers with entailment trees.
\newblock In \emph{Proceedings of the 2021 Conference on Empirical Methods in
  Natural Language Processing}, pp.\  7358--7370, Online and Punta Cana,
  Dominican Republic, November 2021. Association for Computational Linguistics.
\newblock \doi{10.18653/v1/2021.emnlp-main.585}.
\newblock URL \url{https://aclanthology.org/2021.emnlp-main.585}.

\bibitem[Fan et~al.(2019)Fan, Jernite, Perez, Grangier, Weston, and
  Auli]{eli5LQA}
Angela Fan, Yacine Jernite, Ethan Perez, David Grangier, Jason Weston, and
  Michael Auli.
\newblock Eli5: Long form question answering, 2019.
\newblock URL \url{https://arxiv.org/abs/1907.09190}.

\bibitem[Goyal et~al.(2022)Goyal, Li, and Durrett]{goyal-gpt-eval}
Tanya Goyal, Junyi~Jessy Li, and Greg Durrett.
\newblock News summarization and evaluation in the era of gpt-3, 2022.
\newblock URL \url{https://arxiv.org/abs/2209.12356}.

\bibitem[He et~al.(2021)He, Liu, Gao, and Chen]{He2021DeBERTaDB}
Pengcheng He, Xiaodong Liu, Jianfeng Gao, and Weizhu Chen.
\newblock Deberta: Decoding-enhanced bert with disentangled attention.
\newblock \emph{ArXiv}, abs/2006.03654, 2021.

\bibitem[Hessel et~al.(2022)Hessel, Hwang, Park, Zellers, Bhagavatula,
  Rohrbach, Saenko, and Choi]{hessel-sherlock}
Jack Hessel, Jena~D. Hwang, Jae~Sung Park, Rowan Zellers, Chandra Bhagavatula,
  Anna Rohrbach, Kate Saenko, and Yejin Choi.
\newblock The abduction of sherlock holmes: A dataset for visual abductive
  reasoning.
\newblock 2022.
\newblock \doi{10.48550/ARXIV.2202.04800}.
\newblock URL \url{https://arxiv.org/abs/2202.04800}.

\bibitem[Hubbard et~al.(2017)Hubbard, Potts, and Couch]{Hubbard2017HowQT}
Joanna~K. Hubbard, Macy~A. Potts, and Brian~A. Couch.
\newblock How question types reveal student thinking: An experimental
  comparison of multiple-true-false and free-response formats.
\newblock \emph{CBE Life Sciences Education}, 16, 2017.

\bibitem[Inoue et~al.(2020)Inoue, Stenetorp, and Inui]{Inoue2020R4CAB}
Naoya Inoue, Pontus Stenetorp, and Kentaro Inui.
\newblock R4c: A benchmark for evaluating rc systems to get the right answer
  for the right reason.
\newblock In \emph{ACL}, 2020.

\bibitem[Izacard \& Grave(2020)Izacard and Grave]{fid}
Gautier Izacard and Edouard Grave.
\newblock Leveraging passage retrieval with generative models for open domain
  question answering, 2020.
\newblock URL \url{https://arxiv.org/abs/2007.01282}.

\bibitem[Jhamtani \& Clark(2020)Jhamtani and
  Clark]{jhamtani-clark-2020-learning}
Harsh Jhamtani and Peter Clark.
\newblock Learning to explain: Datasets and models for identifying valid
  reasoning chains in multihop question-answering.
\newblock In \emph{Proceedings of the 2020 Conference on Empirical Methods in
  Natural Language Processing (EMNLP)}, pp.\  137--150, Online, November 2020.
  Association for Computational Linguistics.
\newblock \doi{10.18653/v1/2020.emnlp-main.10}.
\newblock URL \url{https://aclanthology.org/2020.emnlp-main.10}.

\bibitem[Karpukhin et~al.(2020)Karpukhin, Oğuz, Min, Lewis, Wu, Edunov, Chen,
  and Yih]{dpr}
Vladimir Karpukhin, Barlas Oğuz, Sewon Min, Patrick Lewis, Ledell Wu, Sergey
  Edunov, Danqi Chen, and Wen-tau Yih.
\newblock Dense passage retrieval for open-domain question answering, 2020.
\newblock URL \url{https://arxiv.org/abs/2004.04906}.

\bibitem[Kingma \& Ba(2014)Kingma and Ba]{kingma2014adam}
Diederik~P Kingma and Jimmy Ba.
\newblock Adam: A method for stochastic optimization.
\newblock \emph{arXiv preprint arXiv:1412.6980}, 2014.

\bibitem[Kwiatkowski et~al.(2019)Kwiatkowski, Palomaki, Redfield, Collins,
  Parikh, Alberti, Epstein, Polosukhin, Devlin, Lee, Toutanova, Jones, Kelcey,
  Chang, Dai, Uszkoreit, Le, and Petrov]{kwiatkowski-etal-2019-natural}
Tom Kwiatkowski, Jennimaria Palomaki, Olivia Redfield, Michael Collins, Ankur
  Parikh, Chris Alberti, Danielle Epstein, Illia Polosukhin, Jacob Devlin,
  Kenton Lee, Kristina Toutanova, Llion Jones, Matthew Kelcey, Ming-Wei Chang,
  Andrew~M. Dai, Jakob Uszkoreit, Quoc Le, and Slav Petrov.
\newblock Natural questions: A benchmark for question answering research.
\newblock \emph{Transactions of the Association for Computational Linguistics},
  7:\penalty0 452--466, 2019.
\newblock \doi{10.1162/tacl_a_00276}.
\newblock URL \url{https://aclanthology.org/Q19-1026}.

\bibitem[Lal et~al.(2021)Lal, Chambers, Mooney, and Balasubramanian]{Lal_2021}
Yash~Kumar Lal, Nathanael Chambers, Raymond Mooney, and Niranjan
  Balasubramanian.
\newblock Tellmewhy: A dataset for answering why-questions in narratives.
\newblock In \emph{Findings of the Association for Computational Linguistics:
  {ACL}-{IJCNLP} 2021}. Association for Computational Linguistics, 2021.
\newblock \doi{10.18653/v1/2021.findings-acl.53}.
\newblock URL \url{https://doi.org/10.18653%2Fv1%2F2021.findings-acl.53}.

\bibitem[Lin(2004)]{lin-2004-rouge}
Chin-Yew Lin.
\newblock {ROUGE}: A package for automatic evaluation of summaries.
\newblock In \emph{Text Summarization Branches Out}, pp.\  74--81, Barcelona,
  Spain, July 2004. Association for Computational Linguistics.
\newblock URL \url{https://aclanthology.org/W04-1013}.

\bibitem[Lin et~al.(2021)Lin, Ma, Lin, Yang, Pradeep, and Nogueira]{pyserini}
Jimmy Lin, Xueguang Ma, Sheng-Chieh Lin, Jheng-Hong Yang, Ronak Pradeep, and
  Rodrigo Nogueira.
\newblock Pyserini: A python toolkit for reproducible information retrieval
  research with sparse and dense representations.
\newblock In \emph{Proceedings of the 44th International ACM SIGIR Conference
  on Research and Development in Information Retrieval}, SIGIR '21, pp.\
  2356–2362, New York, NY, USA, 2021. Association for Computing Machinery.
\newblock ISBN 9781450380379.
\newblock \doi{10.1145/3404835.3463238}.
\newblock URL \url{https://doi.org/10.1145/3404835.3463238}.

\bibitem[Liu et~al.(2019)Liu, Ott, Goyal, Du, Joshi, Chen, Levy, Lewis,
  Zettlemoyer, and Stoyanov]{roberta}
Yinhan Liu, Myle Ott, Naman Goyal, Jingfei Du, Mandar Joshi, Danqi Chen, Omer
  Levy, Mike Lewis, Luke Zettlemoyer, and Veselin Stoyanov.
\newblock Roberta: A robustly optimized bert pretraining approach, 2019.
\newblock URL \url{https://arxiv.org/abs/1907.11692}.

\bibitem[Lu et~al.(2022)Lu, Mishra, Xia, Qiu, Chang, Zhu, Tafjord, Clark, and
  Kalyan]{lu-learntoexplain}
Pan Lu, Swaroop Mishra, Tony Xia, Liang Qiu, Kai-Wei Chang, Song-Chun Zhu,
  Oyvind Tafjord, Peter Clark, and Ashwin Kalyan.
\newblock Learn to explain: Multimodal reasoning via thought chains for science
  question answering, 2022.
\newblock URL \url{https://arxiv.org/abs/2209.09513}.

\bibitem[Mohr et~al.(1989)Mohr, Engelkamp, and Zimmer]{Mohr1989RecallAR}
Gilbert Mohr, Johannes Engelkamp, and Hubert~D. Zimmer.
\newblock Recall and recognition of self-performed acts.
\newblock \emph{Psychological Research}, 51:\penalty0 181--187, 1989.

\bibitem[Pearl(2009)]{pearl2009causality}
Judea Pearl.
\newblock \emph{Causality}.
\newblock Cambridge university press, 2009.

\bibitem[Post(2018)]{post-2018-call}
Matt Post.
\newblock A call for clarity in reporting {BLEU} scores.
\newblock In \emph{Proceedings of the Third Conference on Machine Translation:
  Research Papers}, pp.\  186--191, Belgium, Brussels, October 2018.
  Association for Computational Linguistics.
\newblock URL \url{https://www.aclweb.org/anthology/W18-6319}.

\bibitem[Radford et~al.(2019)Radford, Wu, Child, Luan, Amodei, and
  Sutskever]{Radford2019LanguageMA}
Alec Radford, Jeff Wu, Rewon Child, David Luan, Dario Amodei, and Ilya
  Sutskever.
\newblock Language models are unsupervised multitask learners.
\newblock 2019.

\bibitem[Rajani et~al.(2019)Rajani, McCann, Xiong, and
  Socher]{rajani-etal-2019-explain}
Nazneen~Fatema Rajani, Bryan McCann, Caiming Xiong, and Richard Socher.
\newblock Explain yourself! leveraging language models for commonsense
  reasoning.
\newblock In \emph{Proceedings of the 57th Annual Meeting of the Association
  for Computational Linguistics}, pp.\  4932--4942, Florence, Italy, July 2019.
  Association for Computational Linguistics.
\newblock \doi{10.18653/v1/P19-1487}.
\newblock URL \url{https://aclanthology.org/P19-1487}.

\bibitem[Ribeiro et~al.(2022)Ribeiro, Wang, Ma, Dong, Wei, Zhu, Chen, Huang,
  Xu, Arnold, and Roth]{Ribeiro2022EntailmentTE}
Danilo~Neves Ribeiro, Shen Wang, Xiaofei Ma, Rui Dong, Xiaokai Wei, Henry Zhu,
  Xinchi Chen, Zhiheng Huang, Peng Xu, Andrew~O. Arnold, and Dan Roth.
\newblock Entailment tree explanations via iterative retrieval-generation
  reasoner.
\newblock In \emph{NAACL-HLT}, 2022.

\bibitem[Robertson et~al.(2009)Robertson, Zaragoza,
  et~al.]{robertson2009probabilistic}
Stephen Robertson, Hugo Zaragoza, et~al.
\newblock The probabilistic relevance framework: Bm25 and beyond.
\newblock \emph{Foundations and Trends{\textregistered} in Information
  Retrieval}, 3\penalty0 (4):\penalty0 333--389, 2009.

\bibitem[Sato et~al.(2021)Sato, Yamada, and Kashima]{wmd}
Ryoma Sato, Makoto Yamada, and Hisashi Kashima.
\newblock Re-evaluating word mover's distance, 2021.
\newblock URL \url{https://arxiv.org/abs/2105.14403}.

\bibitem[Sellam et~al.(2020)Sellam, Das, and Parikh]{Sellam2020BLEURTLR}
Thibault Sellam, Dipanjan Das, and Ankur~P. Parikh.
\newblock Bleurt: Learning robust metrics for text generation.
\newblock In \emph{ACL}, 2020.

\bibitem[Shuster et~al.(2022)Shuster, Urbanek, Szlam, and
  Weston]{Shuster2022AmIM}
Kurt Shuster, Jack Urbanek, Arthur~D. Szlam, and Jason Weston.
\newblock Am i me or you? state-of-the-art dialogue models cannot maintain an
  identity.
\newblock In \emph{NAACL-HLT}, 2022.

\bibitem[Sui et~al.(2022)Sui, Feng, Zhang, Cao, Hu, and Zhu]{cf-kgqa}
Yuan Sui, Shanshan Feng, Huaxiang Zhang, Jian Cao, Liang Hu, and Nengjun Zhu.
\newblock Causality-aware enhanced model for multi-hop question answering over
  knowledge graphs.
\newblock \emph{Knowledge-Based Systems}, 250:\penalty0 108943, 2022.
\newblock ISSN 0950-7051.
\newblock \doi{https://doi.org/10.1016/j.knosys.2022.108943}.
\newblock URL
  \url{https://www.sciencedirect.com/science/article/pii/S0950705122004567}.

\bibitem[Talmor et~al.(2019)Talmor, Herzig, Lourie, and
  Berant]{talmor-etal-2019-commonsenseqa}
Alon Talmor, Jonathan Herzig, Nicholas Lourie, and Jonathan Berant.
\newblock {C}ommonsense{QA}: A question answering challenge targeting
  commonsense knowledge.
\newblock In \emph{Proceedings of the 2019 Conference of the North {A}merican
  Chapter of the Association for Computational Linguistics: Human Language
  Technologies, Volume 1 (Long and Short Papers)}, pp.\  4149--4158,
  Minneapolis, Minnesota, June 2019. Association for Computational Linguistics.
\newblock \doi{10.18653/v1/N19-1421}.
\newblock URL \url{https://aclanthology.org/N19-1421}.

\bibitem[Thorne et~al.(2018)Thorne, Vlachos, Christodoulopoulos, and
  Mittal]{thorne-etal-2018-fever}
James Thorne, Andreas Vlachos, Christos Christodoulopoulos, and Arpit Mittal.
\newblock {FEVER}: a large-scale dataset for fact extraction and
  {VER}ification.
\newblock In \emph{Proceedings of the 2018 Conference of the North {A}merican
  Chapter of the Association for Computational Linguistics: Human Language
  Technologies, Volume 1 (Long Papers)}, pp.\  809--819, New Orleans,
  Louisiana, June 2018. Association for Computational Linguistics.
\newblock \doi{10.18653/v1/N18-1074}.
\newblock URL \url{https://aclanthology.org/N18-1074}.

\bibitem[Wadhwa et~al.(2018)Wadhwa, Embar, Grabmair, and
  Nyberg]{Wadhwa2018TowardsIR}
Soumya Wadhwa, Varsha Embar, Matthias Grabmair, and Eric Nyberg.
\newblock Towards inference-oriented reading comprehension: Parallelqa.
\newblock \emph{ArXiv}, abs/1805.03830, 2018.

\bibitem[Wang et~al.(2022)Wang, Roberts, Hesslow, Scao, Chung, Beltagy, Launay,
  and Raffel]{Wang2022WhatLM}
Thomas Wang, Adam Roberts, Daniel Hesslow, Teven~Le Scao, Hyung~Won Chung,
  Iz~Beltagy, Julien Launay, and Colin Raffel.
\newblock What language model architecture and pretraining objective work best
  for zero-shot generalization?
\newblock In \emph{ICML}, 2022.

\bibitem[Wei et~al.(2022)Wei, Wang, Schuurmans, Bosma, Chi, Le, and
  Zhou]{Wei2022ChainOT}
Jason Wei, Xuezhi Wang, Dale Schuurmans, Maarten Bosma, Ed~Chi, Quoc Le, and
  Denny Zhou.
\newblock Chain of thought prompting elicits reasoning in large language
  models.
\newblock \emph{ArXiv}, abs/2201.11903, 2022.

\bibitem[Yang et~al.(2022)Yang, Wang, Wu, Yang, and Zhang]{yang2022towards}
Linyi Yang, Zhen Wang, Yuxiang Wu, Jie Yang, and Yue Zhang.
\newblock Towards fine-grained causal reasoning and qa.
\newblock \emph{arXiv preprint arXiv:2204.07408}, 2022.

\bibitem[Yang et~al.(2018)Yang, Qi, Zhang, Bengio, Cohen, Salakhutdinov, and
  Manning]{yang-etal-2018-hotpotqa}
Zhilin Yang, Peng Qi, Saizheng Zhang, Yoshua Bengio, William Cohen, Ruslan
  Salakhutdinov, and Christopher~D. Manning.
\newblock {H}otpot{QA}: A dataset for diverse, explainable multi-hop question
  answering.
\newblock In \emph{Proceedings of the 2018 Conference on Empirical Methods in
  Natural Language Processing}, pp.\  2369--2380, Brussels, Belgium,
  October-November 2018. Association for Computational Linguistics.
\newblock \doi{10.18653/v1/D18-1259}.
\newblock URL \url{https://aclanthology.org/D18-1259}.

\bibitem[Zaheer et~al.(2020)Zaheer, Guruganesh, Dubey, Ainslie, Alberti,
  Onta{\~n}{\'o}n, Pham, Ravula, Wang, Yang, and Ahmed]{Zaheer2020BigBT}
Manzil Zaheer, Guru Guruganesh, Kumar~Avinava Dubey, Joshua Ainslie, Chris
  Alberti, Santiago Onta{\~n}{\'o}n, Philip Pham, Anirudh Ravula, Qifan Wang,
  Li~Yang, and Amr Ahmed.
\newblock Big bird: Transformers for longer sequences.
\newblock \emph{ArXiv}, abs/2007.14062, 2020.

\bibitem[Zhang et~al.(2020)Zhang, Kishore, Wu, Weinberger, and
  Artzi]{Zhang2020BERTScoreET}
Tianyi Zhang, Varsha Kishore, Felix Wu, Kilian~Q. Weinberger, and Yoav Artzi.
\newblock Bertscore: Evaluating text generation with bert.
\newblock \emph{ArXiv}, abs/1904.09675, 2020.

\end{thebibliography}
\bibliographystyle{iclr2023/iclr2023_conference}
\appendix

\section{Appendix}
\label{sec:appendix}
\subsection{Data Collection}
\label{appendix:collection}
Our corpus consists of the entirety of the English Wikipedia, snapshotted on 23 May 2022. Wikipedia presents a list of curated ``Good Article'', \href{https://en.wikipedia.org/wiki/Wikipedia:Good_articles/all}{Wikipedia's Good articles}, which are articles that are nominated and reviewed to fit the ``Good Article Criteria'', \href{https://en.wikipedia.org/wiki/Wikipedia:Good_article_criteria}{Good article criteria}. Articles from this category are guaranteed to have correct spelling and grammar, as well as clear and concise diction. Our final keyword list includes: ``\texttt{because}'', ``\texttt{due to}'', ``\texttt{therefore}'', ``\texttt{consequently}'', ``\texttt{resulted in}'', ``\texttt{resulting in}'', and ``\texttt{as a result}''.
\subsection{Data Collection Validaton}
\label{appendix:validation}
Each stage in our data collection process is followed by two additional validation layers. 
For Stage 1, workers are prohibited from submitting more than 20 entries until their annotations have been manually validated. 
The annotation result passes through another phase of manual validation to ensure that the quality is kept up after workers’ initial submissions are accepted by quality control. 
For Stage 2, we track a separate list of qualified workers for explanation quality. 

Similar to Stage 1, Stage 2's initial submit limit (the ``speed bump'') is 10.
Undergraduate students manually reviewed the examples from stage-2-qualified workers.
These panelists were instructed and shown demonstrations of marking explanations as satisfying or not and correcting minor errors for slight quality improvements.
While manually approved workers write each \textsc{WikiWhy} explanation, these hand-reviewed samples ultimately comprise the test and development sets.
The continuous flow between stages is enabled by a backend system we implemented to maintain a database of submissions.
This system serves inputs to both MTurk interfaces, as well as the front-end validation interfaces provided to the undergraduate panelists.
\subsection{Additional Results}
\label{appendix:results}
We include additional evaluations of our generated explanations using simple metrics. \autoref{tab:gpt2v3} shows performance on the \textbf{EO} task, and \autoref{tab:gpt_input_tab} show performance on the \textbf{A\&E} task. We also include results from the QA task in \autoref{tab:retrieval} and \autoref{tab:answer}. Automatic evaluation on individual topics categories are included in \autoref{tab:genrescores}.
\subsection{Crowd Worker Interface}
\begin{figure*}
    \centering
    \includegraphics[width=1\linewidth]{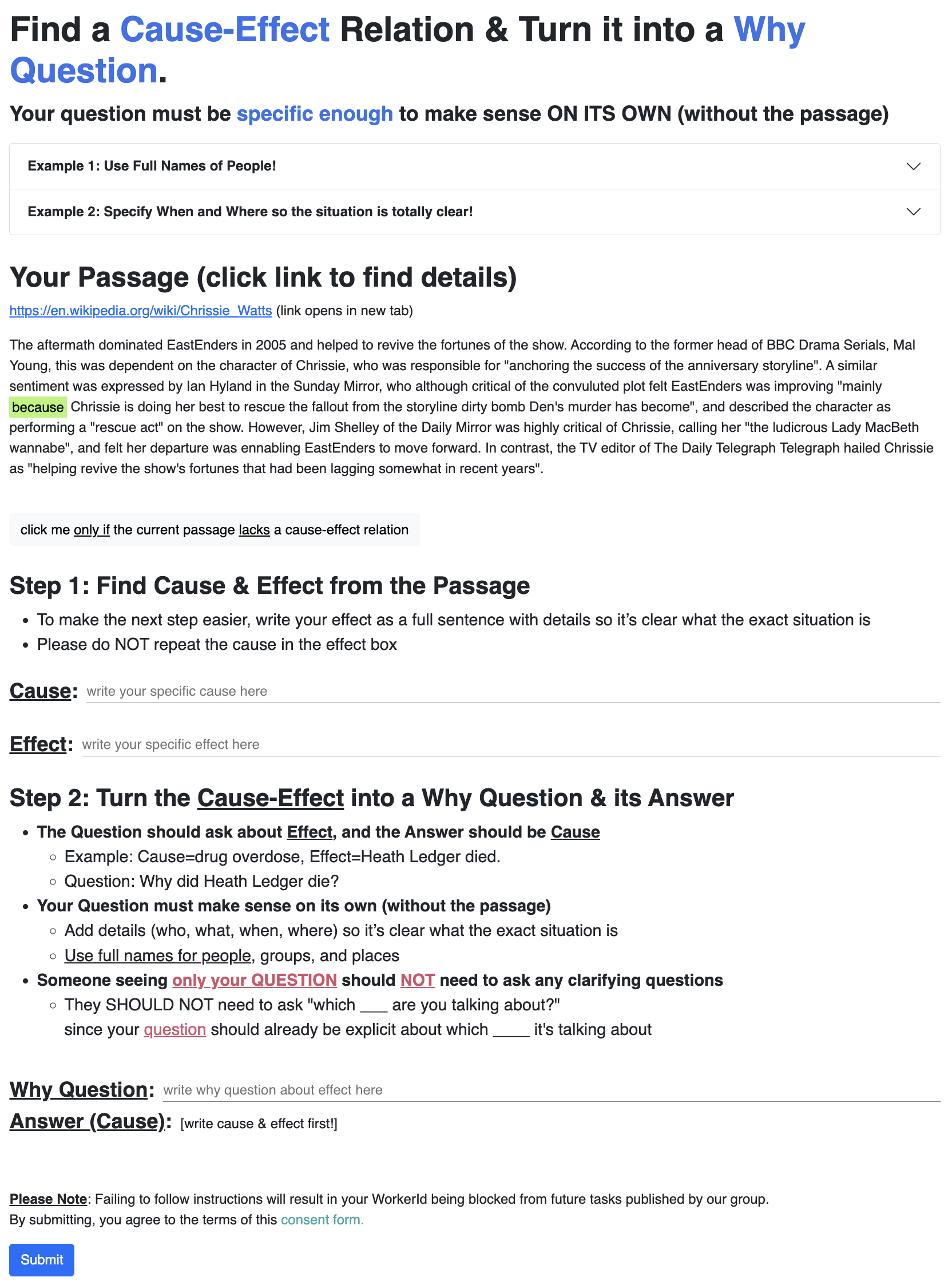}
    \caption{Amazon Mechanical Turk Interface for Stage 1}
    \label{fig:s1interface}
\end{figure*}

\begin{figure*}
    \centering
    \includegraphics[width=1\linewidth]{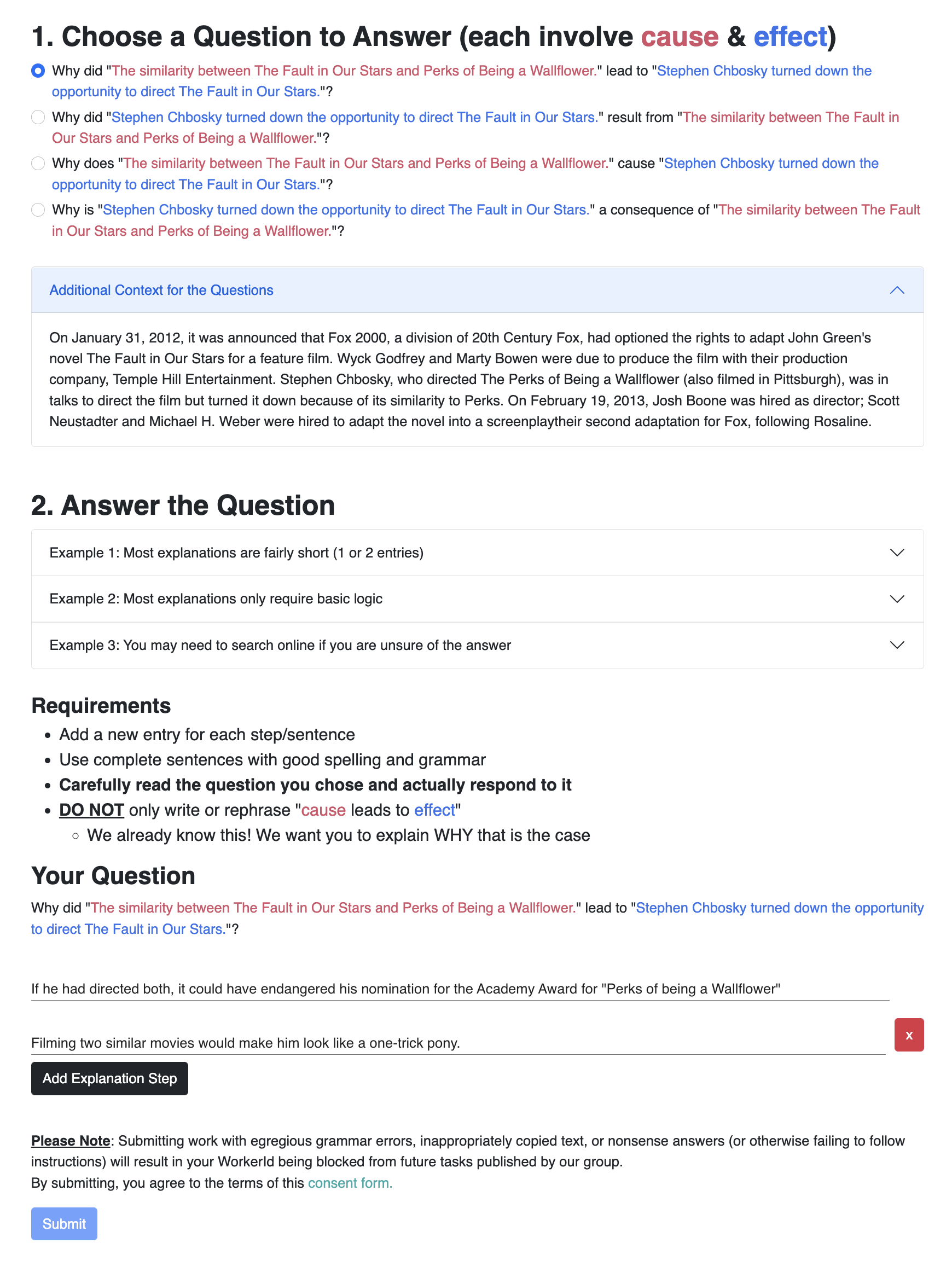}
    \caption{Amazon Mechanical Turk Interface for Stage 2}
    \label{fig:s2interface}
\end{figure*}

\autoref{fig:s1interface} and \autoref{fig:s2interface} display the interfaces for the first and second stages respectively. In addition to the list of requirements, we provide examples and tips to further clarify our expectations. The passage is displayed with a link to the full article so workers can view the complete context if needed. Every passage contains a highlighted causal connective, allowing workers to quickly scan and skip irrelevant portions. Each passage is retrieved from our custom database through our API. If the passage is too difficult for the worker to understand or lacks a cause-effect relation, the worker can click the button below for another random passage.

\newpage
\begin{table*}[t!]
  \centering
    \begin{tabular}{lccccccc}
    \specialrule{.15em}{.05em}{.05em}
     \textbf{Model} & \multicolumn{7}{c}{\textbf{Fine-tuned GPT-2 vs. Few-shot GPT-3}} \\
     \cmidrule{2-8}
     & S-BLEU & WMD & SMS & BERT-f1 & ROUGE-1 & ROUGE-2 & ROUGE-L \\
    \specialrule{.075em}{.05em}{.05em} 
    \textbf{GPT-2} \\
    ~~~\emph{Greedy}  & 0.042 & 0.541 &
    \textbf{15.81} &
    0.773 & 0.212 & \textbf{0.057} &  0.184 \\
    ~~~\emph{Temp 0.5} & 0.037 & 0.540 &  15.30 & 0.770 &   0.198 &  0.047 &  0.169 \\
    ~~~\emph{Temp 1.0} & 0.022 &  0.536 & 13.25 & 0.760 & 0.161 & 0.022 & 0.134 \\
            \specialrule{.075em}{.05em}{.05em} 

    \textbf{GPT-3}\\
    ~~~\emph{Temp 1.0} & \textbf{0.055} & \textbf{0.555} & 14.93 & \textbf{0.792} & \textbf{0.240} & \textbf{0.057} & \textbf{0.199} \\
    \specialrule{.15em}{.05em}{.05em}
    \end{tabular}
    \caption{Explanation Evaluation Results of \textsc{WikiWhy} dataset according to the following metrics: SacreBLEU (S-BLEU) \cite{post-2018-call}, Word-Mover's distance (WMD) \cite{wmd}, Sentence Mover's Similarity Metrics (SMS) \cite{clark-etal-2019-sentence}, BERT-f1 Score \cite{Zhang2020BERTScoreET}, ROUGE-1, ROUGE-2, and ROUGE-L (all ROUGE-f1 Scores \cite{lin-2004-rouge} averaged). SMS is scaled by 1000 for readability.}
  \label{tab:gpt2v3}
\end{table*}

\begin{table*}[t!]
  \centering
    \begin{tabular}{lccccccc}
    \specialrule{.15em}{.05em}{.05em} 
     \textbf{Model} & \multicolumn{7}{c}{\textbf{GPT-3 Prompt Input Experiments}} \\
     \cmidrule{2-8}
     & S-BLEU & WMD & SMS  & BERT-f1 & ROUGE-1  & ROUGE-2 & ROUGE-L \\
    \specialrule{.075em}{.05em}{.05em} 
    \textbf{Input Setting}\\
    ~~~\emph{Ideal} & \textbf{0.055} & \textbf{0.555} & \textbf{14.93} & \textbf{0.792} & \textbf{0.240} &\textbf{ 0.057} & \textbf{0.199} \\
    ~~~\emph{Well-Selected} & 0.030 & 0.546 & 13.27 & 0.776 & 0.203 & 0.049 & 0.149 \\
    ~~~\emph{End-to-end} & 0.023 & 0.542 & 13.22 & 0.768 & 0.200 & 0.038 & 0.144 \\
    \specialrule{.15em}{.05em}{.05em} 
    \end{tabular}
    \caption{\textbf{GPT-3} explanation results with various input settings: Ideal- gold cause/answer, Well-Selected- provided cause/answer predicted by best-performing reader model (\textbf{FiD}), End-to-end- provided only question/effect (\textbf{GPT-3} completes end-to-end task)}
  \label{tab:gpt_input_tab}
\end{table*}

\newpage
\begin{paracol}{2}
    \begin{table}
        \caption[position=top]{\textsc{WikiWhy} dataset contains a diverse set of 11 genres. The raw counts of topic themes in articles is presented in the second column. The relative frequency is the percentage of articles in CausalQA sub-sampled from the \emph{Good} Wikipedia articles list.}
        \centering
        \begin{tabular}{lrc}
            \specialrule{.15em}{.05em}{.05em} 
            \textbf{\textsc{Genres}} & \textbf{\textsc{Raw \#}} & \textbf{\textsc{Freq.}} \\
            \specialrule{.075em}{.05em}{.05em} 
             \textsc{Agriculture} & 131 & 0.436 \\
             \textsc{Arts} & 577 & 0.396 \\
             \textsc{Engineering} & 952 & 0.336 \\
             \textsc{Geography} & 754 & 0.624 \\
             \textsc{History} & 1023 & 0.433 \\
             \textsc{Literature} & 455 & 0.340 \\
             \textsc{Mathematics} & 27 & 0.227 \\
             \textsc{Media} & 1773 & 0.399 \\
             \textsc{Music} & 1070 & 0.229 \\
             \textsc{Natural Sciences} & 2952 & 0.768 \\
             \textsc{Philosophy} & 302 & 0.465 \\
             \specialrule{.15em}{.05em}{.05em} 
        \end{tabular}
    \end{table}
    \switchcolumn
    
    \begin{table}
      \centering
        \begin{tabular}{lcc}
        \specialrule{.15em}{.05em}{.05em} 
        \textsc{Model} & \multicolumn{2}{c}{\textbf{\textsc{WikiWhy}}} \\
        \cmidrule{2-3}
        & {Top-1 Acc} & {MRR} \\
        \specialrule{.075em}{.05em}{.05em} 
         \textbf{BM25} & \textbf{0.810} & \textbf{0.858} \\
         \textbf{DPR} & 0.340 & 0.448 \\
        \specialrule{.15em}{.01em}{.01em} 
        \end{tabular}
                \caption[position=top]{Document Retrieval for \textsc{WikiWhy}. \textbf{BM25} consistently outperforms \textbf{DPR}.}

        \label{tab:retrieval}
    \end{table}

    \begin{table}
  \centering
    \begin{tabular}{lccc}
    \specialrule{.15em}{.05em}{.05em} 
    \textsc{Model} & \multicolumn{3}{c}{\textbf{\textsc{WikiWhy}}} \\
    \cmidrule{2-4}
     & \textbf{S-BLEU} & \textbf{BERT-f1} & \textbf{WMD} \\
     \specialrule{.075em}{.05em}{.05em} 
     \textbf{RoBERTa} \\
     ~~~\emph{Gold} & 0.246 & 0.860 & 0.637 \\
     ~~~\emph{BM25} & 0.214 & 0.832 & 0.620 \\
     \specialrule{.075em}{.05em}{.05em} 
     \textbf{BigBird} \\
     ~~~\emph{Gold} & 0.258 & 0.825 & 0.615 \\
     ~~~\emph{BM25} & 0.223 & 0.802 & 0.602 \\
    \specialrule{.075em}{.05em}{.05em} 
    \textbf{FiD} \\
     ~~~\emph{Gold} & 0.373 & 0.863 & 0.658 \\
     ~~~\emph{BM25} & 0.259 & 0.827 & 0.617 \\
    \specialrule{.15em}{.05em}{.05em} 
    \end{tabular}
    \caption{Answer Evaluation Results for \textsc{WikiWhy} dataset. Stage 1: \textbf{RoBERTa}, \textbf{BigBird}, and \textbf{FiD}. \textbf{FiD Gold} is fine-tuned on 80\% train split \& evaluated on 10\% dev split.}
  \label{tab:answer}
\end{table}
\end{paracol}
\begin{table}[t]
    \centering
    \scriptsize
    \caption{Examples from 6 most frequent topics covered in \textsc{WikiWhy}. $c$ denotes cause, $e$ effect, and $s_i$ the $i$th rationale sentence.}
    \begin{tabular}{ll}
    {\normalsize \textbf{Genres}} & \hspace{5mm} {\normalsize \textbf{Example}} \\
    \specialrule{.1em}{.05em}{.05em} \\ [-1.5ex]
    {\small Geography} & $c$ \hspace{3mm} The geographic isolation of the Hupa homeland \\
    & $s_1$ \hspace{1mm} { The Hupa's homeland was separated by bodies of water or mountains } \\
    & $s_2$ \hspace{1mm} { Not many people could get to the Hupa's homeland } \\
    \vspace{1mm}
    & $e$ \hspace{2mm} { The Hupa had few interactions with early European explorers up to the 19th century} \\ 
    \hline \\ [-1.5ex]
    {\small Literature} & $c$ \hspace{3mm} Increased language contact in the globalizing world \\
    & $s_1$ \hspace{1mm} { Increased contact between people requires increased communication } \\
    & $s_2$ \hspace{1mm} { Speaker of uncommon languages switch to more common languages } \\
    & $s_3$ \hspace{1mm} { Switching away from uncommon languages leads to them being forgotten } \\
    \vspace{1mm}
    & $e$ \hspace{2mm} { Many small languages are becoming endangered as their speakers shift to other languages } \\
    \hline \\ [-1.5ex]
    {\small Media} & $c$ \hspace{3mm} Seeing the Castle of Cagliostro entrenched in Yamazaki that Japan can make high-quality films \\
    & $s_1$ \hspace{1mm} { Viewing The Castle of Cagliostro inspired Takashi Yamazaki } \\
    & $s_2$ \hspace{1mm} { Out of national pride, Takashi Yamazaki followed a model that he believed would produce quality films } \\
    \vspace{1mm}
    & $e$ \hspace{2mm} { Director Takashi Yamazaki modeled his 2019 film Lupin III: The First after The Castle of Cagliostro } \\
    \hline \\ [-1.5ex]
    {\small Music} & $c$ \hspace{3mm} The duration of Hotel California was longer than songs generally played by radio stations \\
    & $s_1$ \hspace{1mm} { Most songs are only 3-4 minutes long } \\
    & $s_2$ \hspace{1mm} { Hotel California is over 6 minutes } \\
    & $s_3$ \hspace{1mm} { People would not want to listen to same song on radio for that long} \\
    \vspace{1mm}
    & $e$ \hspace{2mm} { Don Felder had doubts about the 1997 Eagles song Hotel California } \\
    \hline \\ [-1.5ex]
    {\small Natural Sciences} & $c$ \hspace{3mm} The thermal stress at dawn and dusk \\
    & $s_1$ \hspace{1mm} { The thermal temperatures change so drastically the rocks expand and contract } \\
    & $s_2$ \hspace{1mm} { This process weakens the structural integrity of the rocks } \\
    \vspace{1mm}
    & $e$ \hspace{2mm} { The boulders on Ceres are brittle and degrade rapidly } \\
    \hline \\ [-1.5ex]
    {\small Technology} & $c$ \hspace{3mm} The use of coal power in Turkey \\
    & $s_1$ \hspace{1mm} { Burning coal leads to air pollution } \\
    & $s_2$ \hspace{1mm} { Air pollution causes sickness and early death} \\
    & $s_3$ \hspace{1mm} { Sick and dead people cannot work} \\
    & $e$ \hspace{2mm} { 1.4 million working days were lost across the population of Turkey in 2019 } \\
    \specialrule{.1em}{.05em}{.05em} 
    \end{tabular}
    \label{tab:genre_diversity}
\end{table}
\begin{table}[t]
    \centering
    \caption{Explanation performance (unordered f1) over the most frequent topics. We GPT-2 under the greedy setting and GPT-3 under the same defaults as \autoref{tab:gpt2}}
    \begin{tabular}{lccccccc}
        \specialrule{.15em}{.05em}{.05em}
        & \multicolumn{7}{c}{\textbf{Most Frequent Genres}} \\
        \cmidrule{2-8}
        & ARTS & GEOG & HISTORY & MEDIA & MUSIC & SCIENCE & TECH \\
        \specialrule{.075em}{.05em}{.05em}
        \textbf{Models} \\
        ~~~GPT-2 & 0.256 & 0.221 & 0.202 & 0.161 & 0.239 & 0.252 & 0.236 \\
        ~~~GPT-3 & 0.412 & 0.372 & 0.341 & 0.335 & 0.301 & 0.371 & 0.333 \\
        \specialrule{.15em}{.05em}{.05em}
    \end{tabular}
    \label{tab:genrescores}
\end{table}
\vspace{16pt}
\begin{paracol}{2}
    \begin{table}
        \caption[position=top]{WikiWhy Summary Statistics}
        \centering
        \begin{tabular}{lcc}
            \specialrule{.15em}{.05em}{.05em} 
            \multicolumn{3}{c}{\textbf{WikiWhy Statistics}} \\
                \specialrule{.075em}{.05em}{.05em} 
            \# of Train & & 7,397 \\
            \# of Dev & & 1,004 \\
            \# of Test & & 1,005 \\
            \specialrule{.075em}{.05em}{.05em} 
            \multicolumn{2}{l}{\# of Rationale} & 9,406 \\
            \multicolumn{2}{l}{\# of Rationale Elements} & 14,238 \\
            \multicolumn{2}{l}{Avg. \# Rationale Length} & 1.5137 \\
            \multicolumn{2}{l}{Avg. \# Tokens per Element} & 16.697 \\
            \specialrule{.15em}{.05em}{.05em} 
        \end{tabular}
        \label{tab:summary_stats}
    \end{table}
    \switchcolumn
    \begin{figure}
        \centering
        \includegraphics[width =\columnwidth]{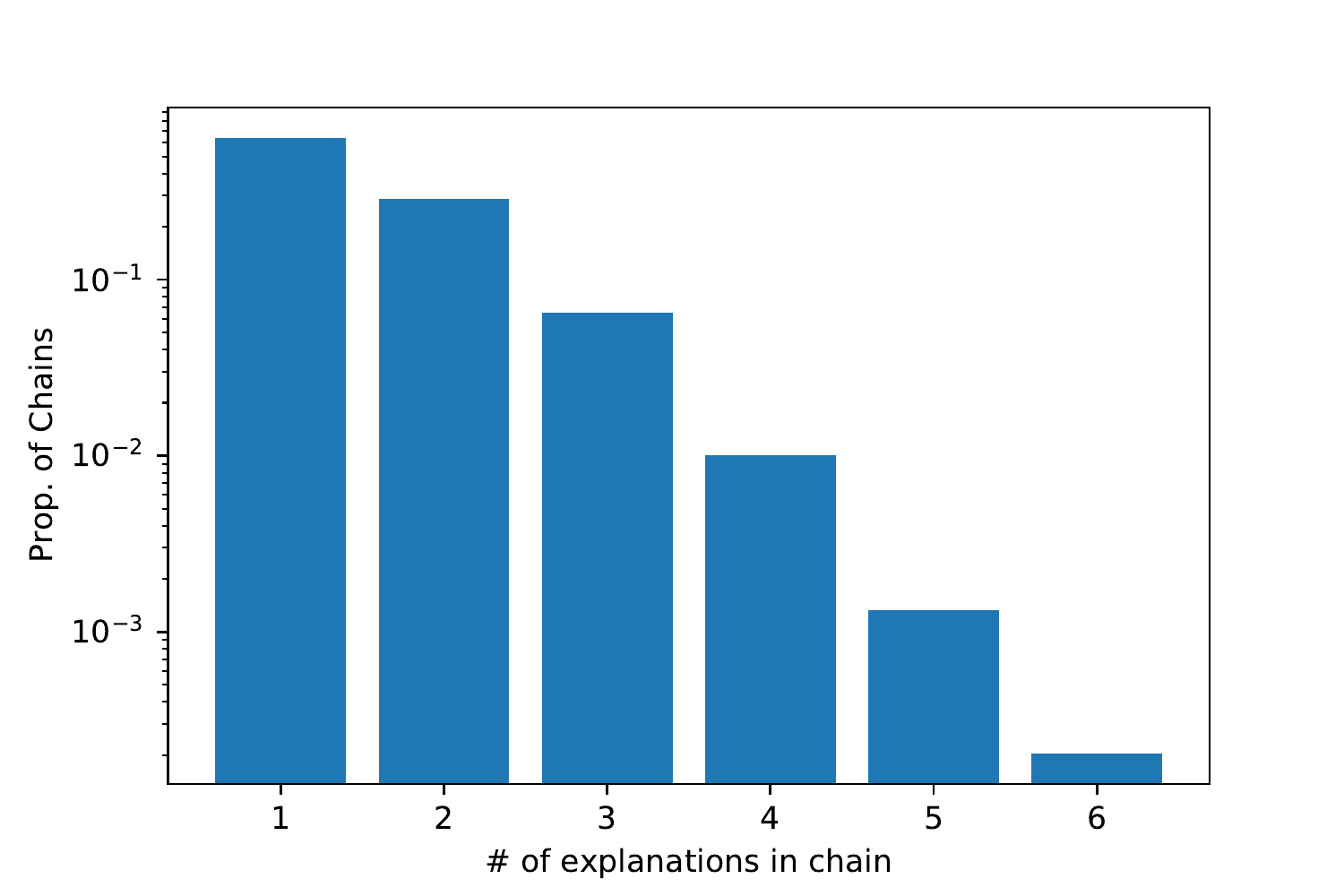}
        \caption{Rationale Length Distribution}
        \label{fig:rationale_length}
    \end{figure}
    \nointerlineskip
\end{paracol}

\begin{figure}
    \centering
    \includegraphics[width=1\linewidth]{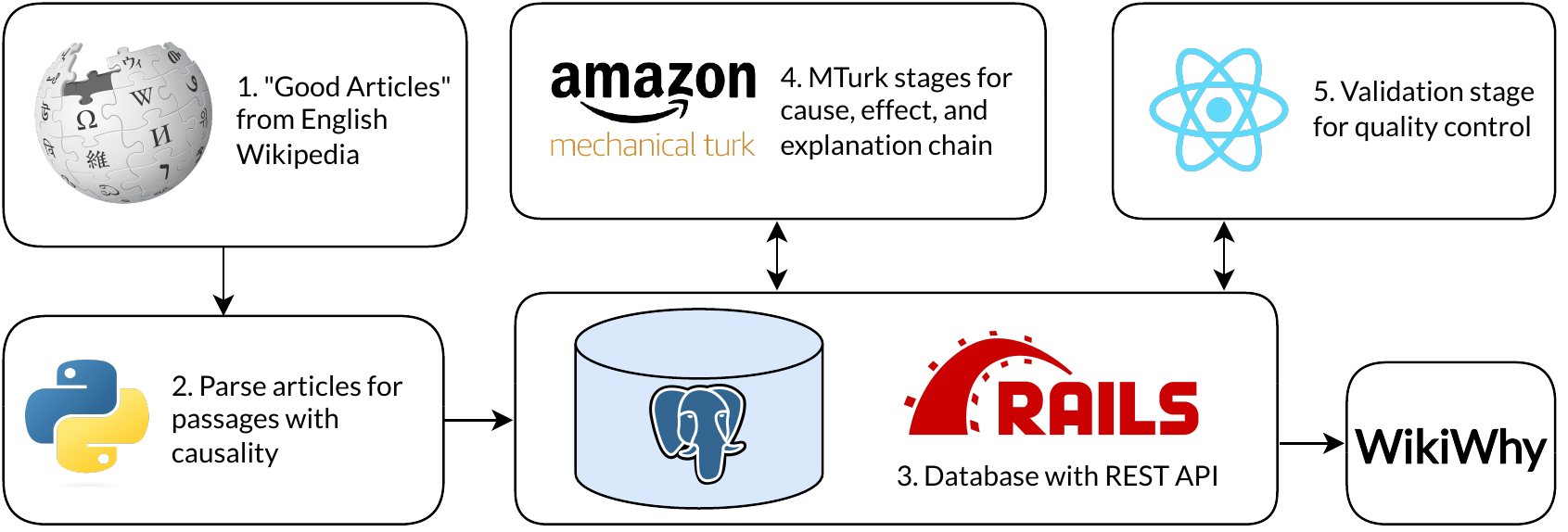}
    \caption{Dataset Collection and Validation Pipeline}
    \label{figure:pipeline}
\end{figure}
\begin{figure*}
    \centering
    \includegraphics[width=1\linewidth]{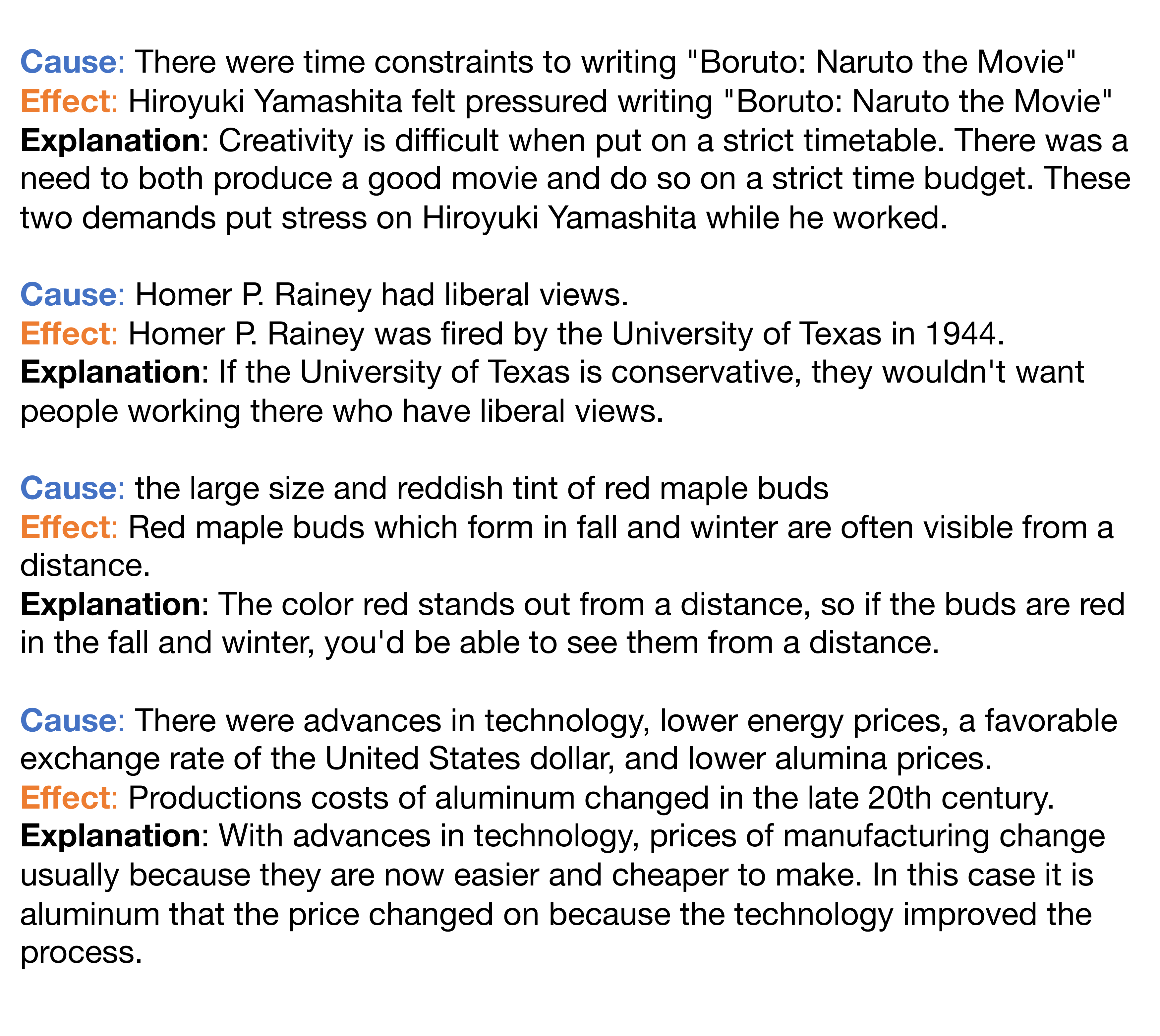}
    \caption{GPT-3 Few-shot Exemplars}
    \label{fig:exemplars}
\end{figure*}

\end{document}